\documentclass[sigconf]{acmart}

\AtBeginDocument{%
  }

\copyrightyear{2024}
\acmYear{2024}
\setcopyright{rightsretained}
\acmConference[CIKM '24]{Proceedings of the 33rd ACM International Conference on Information and Knowledge Management}{October 21--25, 2024}{Boise, ID, USA}
\acmBooktitle{Proceedings of the 33rd ACM International Conference on Information and Knowledge Management (CIKM '24), October 21--25, 2024, Boise, ID, USA}
\acmDOI{10.1145/3627673.3679829}
\acmISBN{979-8-4007-0436-9/24/10}

\usepackage{bbm}
\usepackage{xspace}
\usepackage{xcolor}
\usepackage{subfigure}
\usepackage{algorithm}
\usepackage{algorithmic}

\newcommand{\ourmethod}{SimpDM\xspace}
\definecolor{thered}{RGB}{176,36,24}
\definecolor{theblue}{RGB}{38,56,97}

\newcommand{\sota}[1]{\textbf{#1}}
\newcommand{\runup}[1]{\underline{#1}}

\begin{document}

\title{Self-Supervision Improves Diffusion Models \\for Tabular Data Imputation}

\author{Yixin Liu}
\authornote{This work was completed while the author was an intern at Amazon.}
\email{yixin.liu@monash.edu}
\orcid{0000-0002-4309-5076}
\affiliation{%
  \institution{Monash University}
  \city{Melbourne}
  \country{Australia}
}

\author{Thalaiyasingam Ajanthan}
\authornote{These authors made equal contributions to this work.}
\email{ajthal@amazon.com}
\orcid{0000-0002-6431-0775}
\affiliation{%
  \institution{Amazon}
  \city{Canberra}
  \country{Australia}
}

\author{Hisham Husain}
\authornotemark[2]
\email{hushisha@amazon.com}
\orcid{0009-0007-3736-3410}
\affiliation{%
  \institution{Amazon}
  \city{Melbourne}
  \country{Australia}
}

\author{Vu Nguyen}
\authornotemark[2]
\email{vutngn@amazon.com}
\orcid{0000-0002-0294-4561}
\affiliation{%
  \institution{Amazon}
  \city{Adelaide}
  \country{Australia}
}


\renewcommand{\shortauthors}{Yixin Liu, Thalaiyasingam Ajanthan, Hisham Husain, and Vu Nguyen}

\begin{abstract}
Incomplete tabular datasets are ubiquitous in many applications for a number of reasons such as human error in data collection or privacy considerations. One would expect a natural solution for this is to utilize powerful generative models such as diffusion models, which have demonstrated great potential across image and continuous domains. However, vanilla diffusion models often exhibit sensitivity to initialized noises. This, along with the natural sparsity inherent in the tabular domain, poses challenges for diffusion models, thereby impacting the robustness of these models for data imputation. In this work, we propose an advanced diffusion model named \underline{\textbf{S}}elf-supervised \underline{\textbf{imp}}utation \underline{\textbf{D}}iffusion \underline{\textbf{M}}odel (\ourmethod for brevity), specifically tailored for tabular data imputation tasks. 
To mitigate sensitivity to noise, we introduce a self-supervised alignment mechanism that aims to regularize the model, ensuring consistent and stable imputation predictions. 
Furthermore, we introduce a carefully devised state-dependent data augmentation strategy within \ourmethod, enhancing the robustness of the diffusion model when dealing with limited data. 
Extensive experiments demonstrate that \ourmethod matches or outperforms state-of-the-art imputation methods across various scenarios.

\end{abstract}

\ccsdesc[500]{Information systems~Incomplete data}
\ccsdesc[500]{Computer systems organization~Neural networks}

\keywords{Tabular Data, Data Imputation, Incomplete Data, Diffusion Model, Self-Supervised Learning}


\maketitle

\section{Introduction}
Tabular data are ubiquitous across industries and disciplines, including but not limited to finance~\cite{fin_assefa2020generating}, healthcare~\cite{health_hernandez2022synthetic}, and environmental sciences~\cite{env_judson2009toxicity}. In real-world scenarios, tabular data often contain missing values for many reasons, such as human error, privacy considerations, and the inherent challenges associated with data collection processes~\cite{ot_muzellec2020missing}. 
For instance, some characteristics of a patient might not be accurately documented during their visit~\cite{alaa2017personalized,jarrett2022hyperimpute}. 
The presence of missing data significantly impacts the quality of tabular datasets, introducing bias and rendering a majority of machine learning methods inapplicable.

\begin{figure}
	\centering
	\small
        \subfigure[{With different Gaussian noises (\textcolor{thered}{red}) as initialization, the diffusion model outputs diverse and inaccurate imputation (\textcolor{theblue}{blue}).}]
        {\includegraphics[height=0.35\columnwidth]{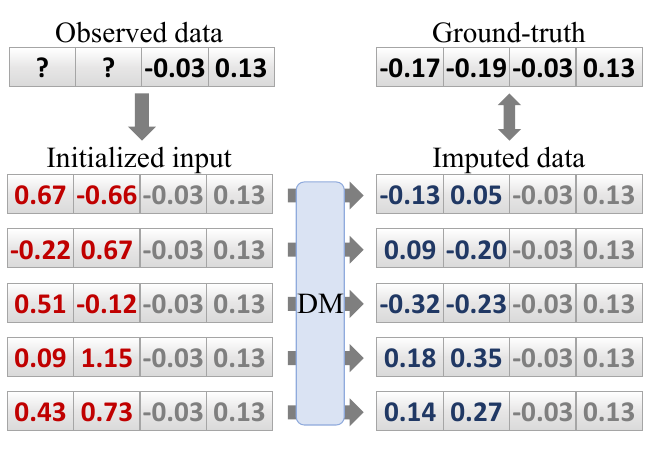}\label{subfig:moti_target}}
        \hfill
	\subfigure[{The limited data samples lead to poor imputation results (higher RMSE) and unstable performance (higher variance).}]{\includegraphics[height=0.35\columnwidth]{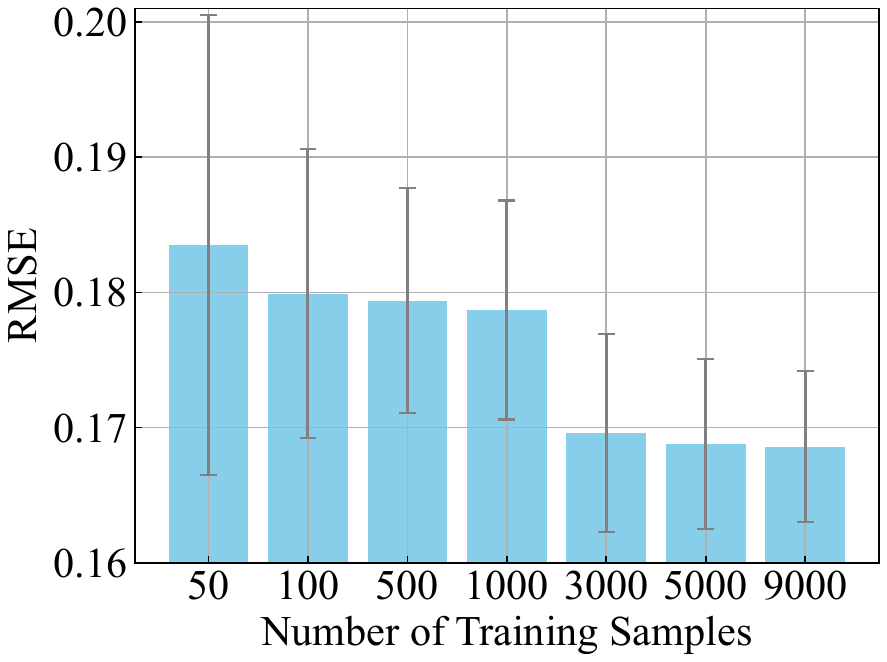}\label{subfig:moti_scale}}
	\caption{Motivating experiments on UCI Power dataset. (a) Given a data sample, the imputation results by diffusion model with different Gaussian initialization at the first diffusion step. (b) The imputation performance under different numbers of training samples.}
\end{figure}

To handle the missing data problem, data imputation is a promising solution that aims to estimate missing values based on the observed data~\cite{survey_osman2018survey}. Existing data imputation methods usually use statistical algorithms~\cite{farhangfar2007novel}, shallow machine learning algorithms~\cite{knnimp_troyanskaya2001missing,mice_van2011mice}, and deep neural networks~\cite{ot_muzellec2020missing,kyono2021miracle} to complete missing data. Among them, deep generative model-based imputation models exhibit competitive performance owing to their capability in modeling the data manifold, which helps complete the missing elements~\cite{yoon2018gain,mattei2019miwae}. 
Presently, diffusion models~\cite{ddpm_ho2020denoising,dmsurvey_yang2022diffusion}, the leading deep generation models, have shown remarkable capabilities in generating data across diverse types, including images~\cite{dmvssurvey_croitoru2023diffusion}, audio~\cite{dmaudio_kong2020diffwave}, and time-series data~\cite{dmts_rasul2021autoregressive}. Building on these achievements, recent studies have used diffusion models for tabular data generation~\cite{kotelnikov2023tabddpm,lee2023codi,kim2022stasy}, as well as data imputation~\cite{tabcsdi_zheng2022diffusion,forestdiff_jolicoeur2023generating}. 

In this research, we identify that vanilla diffusion models are suboptimal for tabular data imputation. 
Specifically, we pinpoint two key mismatches, namely \textbf{objective mismatch} and \textbf{data scale mismatch}, between tabular data imputation tasks and other scenarios where diffusion models excel. These mismatches potentially contribute to the degradation in data imputation performance of diffusion models.

Firstly, the inherent mismatch in \textit{learning objective} between generation and imputation problems can severely impede the imputation performance of diffusion models. Specifically, \textit{diversity} is a key objective of the generation problem, which requires the generated data to vary significantly while maintaining relevance to the given context. 
Diffusion models being sensitive to the initial noise ($x_T$) at the generation stage helps generate diverse samples -- different noise usually leads to different generated samples. 
Conversely, the objective of the imputation task is \textit{accuracy} rather than diversity, requiring the imputed data to closely resemble the singular ground-truth values. In this case, the sensitivity to the initial noise results in a large variance in imputed results, hurting accuracy. 
As shown in the motivating example in Fig.~\ref{subfig:moti_target}, given identical observation ($-0.03$ and $0.13$), the vanilla diffusion model generates diverse imputed values, leading to substantial gaps between imputed values and the ground-truth.

Moreover, the much smaller \textit{data scale} of tabular data compared to other domains (\textit{e.g.}, image) also hinders diffusion models from comprehending the data manifold, yielding subpar data imputation models. For example, CIFAR-10 is a relatively small image dataset, but it still has 60k samples, supporting the diffusion models to capture data patterns~\cite{kulikov2023sinddm}. In contrast, tabular data usually have only a few thousand or even hundreds of samples, making it much more difficult for diffusion models to capture the true data distribution, resulting in overfitting. An example in Fig.~\ref{subfig:moti_scale} shows the negative impact of data samples on imputation effectiveness.

To improve the performance of diffusion models on tabular data imputation tasks, we introduce an advanced diffusion model termed \underline{\textbf{S}}elf-supervised \underline{\textbf{imp}}utation \underline{\textbf{D}}iffusion \underline{\textbf{M}}odel (\ourmethod for short). 
By integrating self-supervised learning techniques into tabular diffusion model, we address the aforementioned mismatch issues. 
Specifically, to tackle objective mismatch, we introduce a self-supervised alignment mechanism to regularize the output of the diffusion model. The key idea is to encourage the diffusion model to provide consistent and accurate imputation for the same observed data, enhancing the stability of imputation results. 
Furthermore, to handle data scale mismatch, we introduce state-dependent augmentation, a perturbation-based data augmentation strategy that is carefully designed for tabular data imputation tasks. With data augmentation, we extend the training set, improving the robustness of the diffusion model. Meanwhile, an effective augmentation can also ensure the effectiveness of the self-supervised learning framework. 
To verify the imputation capability of our \ourmethod model, we conducted extensive experiments on real-world benchmark datasets across multiple missing data scenarios. The empirical results highlight the strong imputation performance of \ourmethod compared to state-of-the-art methods. 
To sum up, the contributions of this paper are three-fold:

\begin{itemize}
    \item \textbf{Finding.} We identify two key challenges that hinder diffusion models in solving tabular data imputation tasks: target mismatch and data scale mismatch.
    \item \textbf{Solution.} We propose a novel diffusion model termed \ourmethod. The proposed method integrates two effective techniques, i.e., self-supervised alignment and state-dependent augmentation, to address the aforementioned challenges.
    \item \textbf{Experiments.} We conduct extensive experiments on 17 benchmark datasets in multiple missing scenarios, and the results illustrate the effectiveness of \ourmethod.
\end{itemize}

\section{Related Works}
In this section, we briefly review two related research directions, i.e., tabular missing data imputation and diffusion models. 

\subsection{Tabular Missing Data Imputation}

Tabular data imputation is an essential research topic to handle the missing data problem. Early solutions use statistical algorithms to estimate missing entries according to the mean, median, or mode estimation of observed data~\cite{farhangfar2007novel}. Besides statistical methods, shallow machine learning methods such as kNN imputation~\cite{knnimp_troyanskaya2001missing}, MICE~\cite{mice_van2011mice}, and missForest~\cite{stekhoven2012missforest} are also effective to complete the missing data. 

To further exploit the potential of deep learning techniques, recent studies use deep neural networks for data imputation ~\cite{jarrett2022hyperimpute}. For instance, Muzellec et al.~\shortcite{ot_muzellec2020missing} propose training deep imputation model by minimizing the optimal transport distance between two groups of data, while Kyono et al.~\shortcite{kyono2021miracle} use deep models to discover the causal structure underlying data for data imputation. Graph neural networks~\cite{liu2024arc,liu2023dink,zheng2023finding,liang2024mines,zheng2023towards} are also adapted to data imputation since they are capable of modeling inter-sample correlation~\cite{grape_you2020handling,telyatnikov2023egg,igrm_zhong2023data}. 
Within deep methodologies, generative model-based approaches stand out for their remarkable performance. Specifically, GAIN~\cite{yoon2018gain} harnesses GAN while MIWAE~\cite{mattei2019miwae} utilizes VAE as their backbone, enhancing their imputation capabilities. Several recent studies also attempted to apply diffusion models to tabular data imputation tasks~\cite{tabcsdi_zheng2022diffusion,forestdiff_jolicoeur2023generating,ouyang2023missdiff}. Nevertheless, these approaches tend to underestimate the inherent disparity between data imputation and diffusion models, resulting in suboptimal imputation performance. %

\subsection{Diffusion Models}

Diffusion models~\cite{sohl2015deep} are a generative paradigm that strives to approximate the target distribution through the endpoint of a Markov chain. This chain initiates from a specified parametric distribution, usually a standard Gaussian. Each step of this Markov process is executed by a deep neural network that learns the inversion of the diffusion process. DDPM~\cite{ddpm_ho2020denoising} bridges the gap between diffusion models and score matching approaches~\cite{song2020improved}, showcasing the powerful capability of diffusion models in image generation. So far, diffusion models have made great success in the generation tasks on various domains, such as image~\cite{dhariwal2021diffusion}, text~\cite{multinomial_hoogeboom2021argmax}, audio~\cite{dmaudio_kong2020diffwave}, time-series~\cite{dmts_rasul2021autoregressive}, and graph~\cite{graphdm_xu2022geodiff,pan2024integrating}. 

Given the formidable capabilities of diffusion models, recent studies attempt to leverage diffusion models to handle learning tasks on tabular data. For instance, TabDDPM~\cite{kotelnikov2023tabddpm} is a representative method that combines Gaussian and Multinomial diffusion models together to generate mixed-type tabular data. CoDi~\cite{lee2023codi} and STaSy~\cite{kim2022stasy} also show impressive capability in tabular data synthesis. Inspired by the success of diffusion models in image inpainting~\cite{lugmayr2022repaint} and time-series imputation~\cite{tashiro2021csdi}, researchers start to discover the potential of diffusion models in tabular data imputation~\cite{tabcsdi_zheng2022diffusion,forestdiff_jolicoeur2023generating,ouyang2023missdiff}. This paper further delves into this research direction, enhancing diffusion models for imputation from a novel perspective.

\section{Preliminary}
\subsection{Problem Definition}

Let $\overline{\mathbf{X}} = \{\overline{\mathbf{x}}_{[i]}\}^n_{i=1}$ denotes the feature matrix of a complete tabular dataset, where the $i$-th row $\overline{\mathbf{x}}_{[i]}$ is a $d$-dimensional feature vector of the $i$-th sample. Each feature can be numerical or categorical. The missing data problem %
can be modeled by a binary mask matrix $\mathbf{M} \in \{0,1\}^{n\times d}$, where $\mathbf{M}_{[i,j]}=1$ indicates the entry $\overline{\mathbf{X}}_{[i,j]}$ is missing, otherwise $\mathbf{M}_{[i,j]}=0$. The observed incomplete data matrix can be represented by
\begin{equation}
\mathbf{X}=\mathbf{X}^{(obs)} \odot \left(\mathbbm{1}_{n \times d}-\mathbf{M}\right)+ \varnothing_{n \times d} \odot \mathbf{M},
\end{equation}

\noindent where $\varnothing_{n \times d}$ is an $n \times d$ matrix of null (missing) values, $\mathbf{X}^{(obs)}$ contains the observed entries that are from $\overline{\mathbf{X}}$, $\odot$ is the element-wise product and $\mathbbm{1}_{n \times d}$ is an $n \times d$ matrix containing $1$ for every entry. Given $\mathbf{X}$, the goal of \textbf{tabular data imputation} is to estimate an imputed data matrix $\hat{\mathbf{X}}$ where the missing entries of $\mathbf{X}$ are filled, which can be written by
\begin{equation}
\hat{\mathbf{X}}=\mathbf{X}^{(obs)} \odot \left(\mathbbm{1}_{n \times d}-\mathbf{M}\right)+\hat{\mathbf{X}}^{(imp)} \odot \mathbf{M},
\end{equation}

\noindent where $\hat{\mathbf{X}}^{(imp)}$ contains the imputed entries. The objective is to learn $\hat{\mathbf{X}}$ that is as close as possible to $\overline{\mathbf{X}}$.

\subsection{Diffusion Models}

Diffusion models~\cite{sohl2015deep} are deep generative models derived from a forward and reverse Markov process. The forward process $q$ is to gradually disturb a sample $\mathbf{x}_0$ into a noisy sample $\mathbf{x}_T$, while the reverse process $p$ is to denoise and generate the sample $\hat{\mathbf{x}}_0$ from $\mathbf{x}_T$:
\begin{equation}
\begin{aligned}
&q\left(\mathbf{x}_{1: T} \mid \mathbf{x}_0\right):=\prod_{t=1}^T q\left(\mathbf{x}_t \mid \mathbf{x}_{t-1}\right), \\
&p_\theta\left(\mathbf{x}_{0: T}\right):=p\left(\mathbf{x}_T\right) \prod_{t=1}^T p_\theta\left(\mathbf{x}_{t-1} \mid \mathbf{x}_t\right),
\end{aligned}
\end{equation}
\noindent where $p_\theta\left(\mathbf{x}_{t-1} \mid \mathbf{x}_t\right)$ is parametrized by a neural network whose parameter is $\theta$. Here $\theta$ can be optimized by minimizing the variational upper bound on the negative log-likelihood:
\begin{equation}
\begin{aligned}
\mathcal{L}_{\mathrm{vb}} & =\mathbb{E}_q[\underbrace{D_{\mathrm{KL}}\left[q\left(\mathbf{x}_T \mid \mathbf{x}_0\right)|| p\left(\mathbf{x}_T\right)\right]}_{L_T} \underbrace{-\log p_\theta\left(\mathbf{x}_0 \mid \mathbf{x}_1\right)}_{L_0} \\
& +\sum_{t=2}^T \underbrace{D_{\mathrm{KL}}\left(q\left(\mathbf{x}_{t-1} \mid \mathbf{x}_t, \mathbf{x}_0\right)|| p_\theta\left(\mathbf{x}_{t-1} \mid \mathbf{x}_t\right)\right)}_{L_{t-1}}] .
\end{aligned}
\end{equation}

We can describe the evolution of the distributions with the following forward and reverse process equations
\begin{align}
    &d\mathbf{x}_t = \mathbf{x}_t dt + \sqrt{2} d\mathbf{w},\\
    &d\mathbf{x}_t = \left (\mathbf{x}_t - 2 \nabla_x \log p_t \right) + \sqrt{2} d\mathbf{w},
\end{align}
where $\mathbf{w}$ corresponds to Brownian motion and $p_t$ is the law of the random variable $\mathbf{x}_t$. Gaussian diffusion models~\cite{ddpm_ho2020denoising} have forward and reverse processes characterized by Gaussian distributions:
\begin{equation}
\begin{aligned}
& q\left(\mathbf{x}_t \mid \mathbf{x}_{t-1}\right)=\mathcal{N}\left(\mathbf{x}_t ; \sqrt{1-\beta_t} \mathbf{x}_{t-1}, \beta_t I\right), \\
& q\left(\mathbf{x}_T\right)=\mathcal{N}\left(\mathbf{x}_T ; 0, I\right), \\
& p_\theta\left(\mathbf{x}_{t-1} \mid \mathbf{x}_t\right)=\mathcal{N}\left(x_{t-1} ; \mu_\theta\left(\mathbf{x}_t, t\right), \Sigma_\theta\left(\mathbf{x}_t, t\right)\right),
\end{aligned}
\end{equation}

\noindent where Gaussian noise is added to the sample following a variance schedule $\beta_t \in (0,1)$. Using equivalences between score matching and error matching in \cite{hyvarinen2005estimation}, \cite{ddpm_ho2020denoising} further propose a simplified objective function for model optimization:
\begin{equation}
\label{eq:dmloss}
\mathcal{L}_{\text {ddpm }}=\mathbb{E}_{t, \mathbf{x}_0, \boldsymbol{\epsilon}}\left[\left\|\boldsymbol{\epsilon}-\boldsymbol{\epsilon}_\theta\left(\mathbf{x}_t, t\right)\right\|^2\right],
\end{equation}

\noindent where $\boldsymbol{\epsilon}_\theta$ is a neural network aiming to predict the Gaussian noise $\boldsymbol{\epsilon}$, which can be used to generate $\hat{\mathbf{x}}_0$ during inference.

To deal with categorical data, multinomial diffusion models~\cite{multinomial_hoogeboom2021argmax} define a categorical distribution that perturbs the data with noise over $K$ classes:
\begin{equation}
\begin{aligned}
&q\left(\mathbf{x}_t \mid \mathbf{x}_{t-1}\right)=\operatorname{Cat}\left(\mathbf{x}_t ;\left(1-\beta_t\right) \mathbf{x}_{t-1}+\beta_t / K\right),\\
&p_\theta\left(\mathbf{x}_{t-1} \mid \mathbf{x}_t\right)=\sum^K_{\hat{\mathbf{x}}_0=1} q\left(\mathbf{x}_{t-1} \mid \mathbf{x}_t, \hat{\mathbf{x}}_0\right) p_\theta\left(\hat{\mathbf{x}}_0 \mid \mathbf{x}_t\right),
\end{aligned}
\end{equation}

\noindent where $\operatorname{Cat}(\cdot)$ is a categorical distribution.

\section{Methodology}
\begin{figure*}[tb]
	\centering
	\includegraphics[width=1\textwidth]{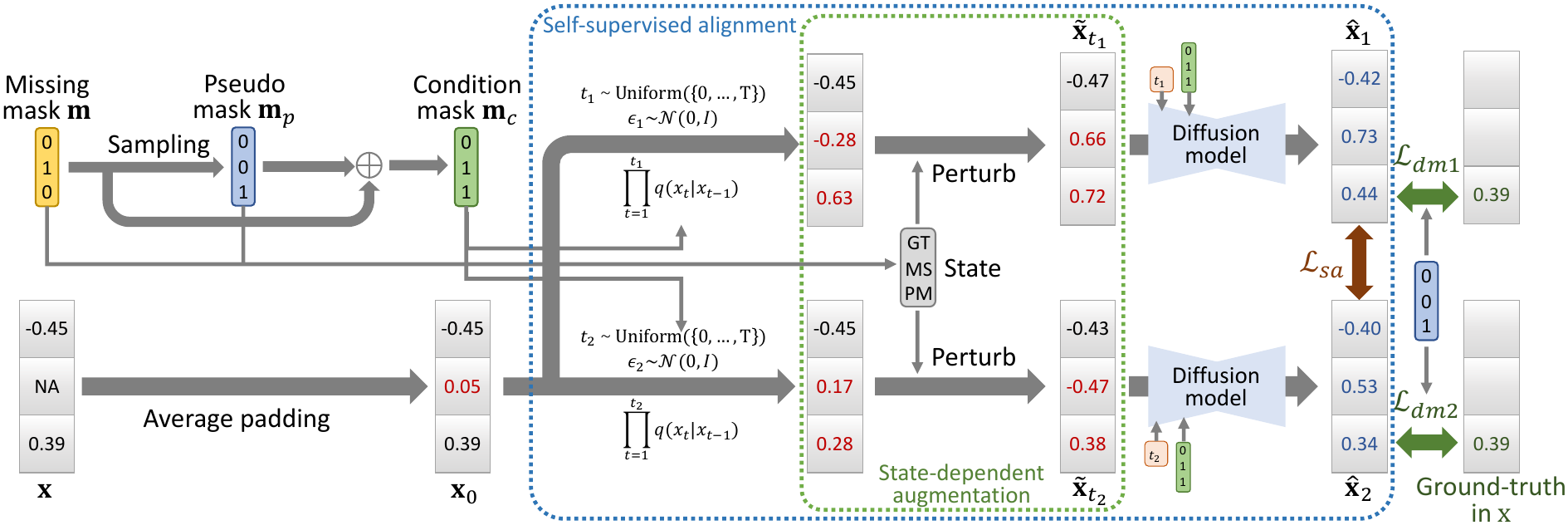} 
	\caption{The overall pipeline of the training procedure of \ourmethod. Given a training sample $\mathbf{x}$ and its missing mask $\mathbf{m}$, the first step is to apply average padding for the missing entries and sample the pseudo mask $\mathbf{m}_p$ and condition mask $\mathbf{m}_c$. In self-supervised alignment, we sample different $t$ and $\epsilon$ at two channels, and then run the diffusion model at each channel. Apart from the diffusion model loss $\mathcal{L}_{dm}$, we use a self-supervised alignment loss $\mathcal{L}_{sa}$ to minimize the distance of the predictions at two channels. We further use a state-dependent augmentation strategy to perturb the model's input according to the states (GT, MS, or PM) of each entry.}
	\label{fig:pipeline}
\end{figure*} 

In this section, we present \ourmethod, a self-supervision improved diffusion model for tabular data imputation. The overall training pipeline of \ourmethod is demonstrated in Fig.~\ref{fig:pipeline}. In Sec.~\ref{subsec:base_model}, we first introduce the base diffusion model specially constructed for tabular data imputation. We then extend the base model to \ourmethod by introducing two pivot designs, \textit{i.e.}, state-dependent self-supervised alignment (Sec.~\ref{subsec:ssl}) and data augmentation (Sec.~\ref{subsec:aug}). Finally, we discuss how to apply \ourmethod to mixed-type tabular data in Sec.~\ref{subsec:mix}. 

\subsection{Diffusion Model for Data Imputation}\label{subsec:base_model}

We first introduce a Gaussian diffusion model for imputing tabular data with numerical features. Note that the vanilla diffusion models~\cite{ddpm_ho2020denoising} are mainly designed for data generation,  making it difficult to apply them directly to imputation tasks~\cite{lugmayr2022repaint}. Therefore, we implement a series of modifications to adapt the diffusion model effectively for tabular data imputation. 

\noindent\textbf{Hybrid input data.}
A major characteristic of the data imputation problem is that the data is partly accessible. In other words, we have the observed entries to guide the estimation of the missing ones. To involve the observed entries in model input, we use a ``hybrid input data'' design in \ourmethod. More precisely, the model input consists of both the ground-truth values from observed entries and the initialized/estimated values for the missing entries, combined together. Such a design, similar to RePaint~\cite{lugmayr2022repaint} for image inpainting, allows the diffusion model to effectively leverage known information to infer missing data while preserving consistency across various samples in the input space.

\noindent\textbf{MLP-based diffusion model.}
As tabular data is usually low-dimensional and with few data samples, complex network model architectures are prone to overfitting and rather unnecessary. Since previous study~\cite{kotelnikov2023tabddpm} shows that a shallow MLP-based diffusion model is adequate for tabular data generation, in \ourmethod, we use a plain MLP as our model, without U-net architecture and cross-layer shortcut. The simple model architecture not only substantially reduces the computational cost of our method but also guards against overfitting the limited data. 
The learnable embedding of diffusion step $t$ is added to the latent representation at the first layer. Notably, we further introduce the \textit{missing mask as an extra input condition} of the diffusion model. In particular, we use a learnable projection layer to map the missing mask vector to a mask embedding, and then add the mask embedding into the first-layer latent representation. The mask embedding enables the diffusion model to discern the status of the hybrid input, aiding in making accurate imputations. 

\noindent\textbf{Pseudo missing training strategy.}
In the setting of data imputation problem, a key challenge is that the ground-truth values of missing entries are unknown during training procedure~\cite{kyono2021miracle}. That is to say, we cannot acquire supervision signals from these entries for model training. To address this challenge, we develop a ``pseudo missing'' training strategy inspired by CSDI~\cite{tashiro2021csdi}. Firstly, during the training phase, we pad the values of missing entries to their respective column-wise average values, which can be regarded as an estimation via mean imputation. Then, in each training iteration, we sample a pseudo missing mask to remask a partition of observed entries. Subsequently, we merge the original missing mask and pseudo mask to create the condition mask of the current training iteration. For the entries where the condition mask equals $1$, we add Gaussian noise $\epsilon$ to them based on the sampled diffusion time step $t$. As for the remaining entries, which act as ``observed data'' in the current iteration, we maintain their original values. In this case, we can train the model by denoising the pseudo missing entries since their ground-truth values are accessible. 

\noindent\textbf{Value prediction and loss calculation.}
In vanilla Gaussian diffusion models~\cite{ddpm_ho2020denoising}, predicting the Gaussian noises of diffusion (Eq.~(\ref{eq:dmloss})) has been shown to be effective for data generation. However, in the context of tabular data imputation, our empirical findings suggest that predicting the missing values can result in improved imputation performance compared to predicting noises. Hence, we directly let the model predict the ground-truth values of the missing entries. When computing the loss function, we solely calculate the loss on the pseudo missing terms, directing the model's attention toward completing the missing data. 

To sum up, the training process of Gaussian diffusion model for tabular data imputation can be expressed by:
\begin{equation}
\label{eq:ourdmloss}
\mathcal{L}_{\text {dm}}=\mathbb{E}_{t, \mathbf{x}_0, \boldsymbol{\epsilon},\mathbf{m}_{p}}\left[\left\|(\mathbf{x}_0-\mathbf{x}_\theta\left(\tilde{\mathbf{x}}_t, t, \mathbf{m}_{c}\right))\odot \mathbf{m}_{p}\right\|^2\right],
\end{equation}
\begin{equation}
\label{eq:hybrid}
\tilde{\mathbf{x}}_t = {\mathbf{x}}_0 \odot \left(\mathbbm{1}_{d}-\mathbf{m}_c\right) + \mathbf{x}_t \odot \mathbf{m}_c,
\end{equation}
\noindent where $\mathbf{m}_p$ and $\mathbf{m}_c = \mathbf{m} + \mathbf{m}_p$ are the pseudo mask and condition mask respectively, $\tilde{\mathbf{x}}_t$ is the hybrid model input, and $\mathbf{x}_\theta$ is the MLP-based model. During inference time, we can initialize the missing entries by $q(\mathbf{x}_T)$ and set condition mask $\mathbf{m}_c =\mathbf{m}$ directly. Before each inference denoising step, we execute Eq.~(\ref{eq:hybrid}) to ensure the observed data is embedded in the model input. 

\subsection{Self-Supervised Alignment}\label{subsec:ssl}

With the diffusion model introduced in Sec.~\ref{subsec:base_model} (we denote it as ``base model'' in the rest of this paper), we can complete missing entries with recurrent denoising processes. Nevertheless, the base model struggles to achieve optimal imputation performance due to the inherent target mismatch between imputation and generation tasks. Specifically, to achieve the diversity target in generation tasks, diffusion models are sensitive to the initial noise $x_T$. With varying initial noises, the generated data should exhibit diversity. However, this sensitivity contrasts with the goals of data imputation tasks, which require precise prediction of missing values rather than diversity. To further improve the imputation performance, we design a self-supervised alignment mechanism to suppress this sensitivity.

Concretely, in the training procedure of \ourmethod, we construct two parallel channels to run the diffusion model for each sample. The two channels share the same pseudo mask and condition mask. For each channel, we sample its diffusion step (denoted by $t_1$ and $t_2$) and diffusion noise (denoted by $\epsilon_1$ and $\epsilon_2$), respectively. With different $t$ and $\epsilon$, we can execute the forward diffusion process to generate $x_{t_1}$ and $x_{t_2}$, and finally acquire the corresponding hybrid inputs $\tilde{x}_{t_1}$ and $\tilde{x}_{t_2}$ via Eq.~(\ref{eq:hybrid}). Since they have a shared condition mask, the observed data in $\tilde{x}_{t_1}$ and $\tilde{x}_{t_2}$ are identical. As the example shown in Fig.~\ref{fig:pipeline}, both of the two channels have an observed value $-0.45$. Differently, due to the disparity between $x_{t_1}$ and $x_{t_2}$, the pseudo missing and real missing entries in $\tilde{x}_{t_1}$ and $\tilde{x}_{t_2}$ exhibit notable differences. 

Recalling that our goal is to suppress the sensitivity to diverse noisy inputs. In particular, given two input data with the same observed entries (i.e., $\tilde{x}_{t_1}$ and $\tilde{x}_{t_2}$), the outputs of the diffusion model should be close to each other. Motivated by this, we acquire their corresponding outputs $\hat{\mathbf{x}}_1=\mathbf{x}_\theta\left(\tilde{\mathbf{x}}_{t_1}, {t_1}, \mathbf{m}_{c}\right)$ and $\hat{\mathbf{x}}_2=\mathbf{x}_\theta\left(\tilde{\mathbf{x}}_{t_2}, {t_2}, \mathbf{m}_{c}\right)$, and try to minimize the difference between them with a self-supervised alignment loss $\mathcal{L}_{sa}(\hat{\mathbf{x}}_1,\hat{\mathbf{x}}_2)$. At the same time, we calculate the basic loss of the diffusion model (Eq.~(\ref{eq:ourdmloss})) at two channels, writing the final loss function of \ourmethod as:
\begin{equation}
\label{eq:overall_loss}
\mathcal{L}=\mathcal{L}_{dm1}(\hat{\mathbf{x}}_1) + \mathcal{L}_{dm2}(\hat{\mathbf{x}}_2) + \gamma \mathcal{L}_{sa}(\hat{\mathbf{x}}_1,\hat{\mathbf{x}}_2),
\end{equation}

\noindent where $\gamma$ is a tunable trade-off hyper-parameter for $\mathcal{L}_{sa}$. In practice, we have several options for $\mathcal{L}_{sa}$, such as MSE loss, contrastive loss~\cite{simclr}, and Sinkhorn divergence~\cite{ot_muzellec2020missing}. Considering its empirical performance (see Sec.~\ref{subsec:ablation}) and high efficiency, we employ MSE loss for self-supervised alignment in \ourmethod.

\subsection{State-Dependent Data Augmentation}\label{subsec:aug}

Another significant challenge in tabular data imputation is the limited size of the dataset, which might not offer adequate information for diffusion models to learn the data manifold effectively. Consequently, the model can lean towards overfitting due to data scarcity, diminishing the overall robustness of diffusion models. To address this problem, data augmentation emerges as a promising solution, creating extra synthetic samples from the original ones~\cite{aug_shorten2019survey,kulikov2023sinddm}. Despite various data augmentation methods designed for image data, most of them cannot be applied to tabular data, which motivates us to produce a well-crafted augmentation strategy for our target scenario. %

To augment tabular data, data perturbation with random noises (\textit{e.g.}, uniform, Gaussian, or others) can be a simple yet effective strategy~\cite{sathianarayanan2022feature,cai2022plad,wang2024unifying}. Nevertheless, this simple strategy may not fully adapt to the diffusion model for \ourmethod where input entries are missing or already noisy. Specifically, weak perturbations might minimally affect the missing or uncertain input entries since they are already noisy. However, increasing the perturbation strength significantly shifts the augmented data away from the original data distribution.

\begin{table*}[th]
    \caption{Imputation performance comparison in terms of RMSE. The best and runner-up performances are highlighted by \textbf{bold} and \underline{underline}, respectively. ``OOM'' indicates Out-Of-Memory on a 16GB GPU.}
    \centering
    \resizebox{1.0\textwidth}{!}{
    \begin{tabular}{l|ccccccccc}
        \toprule
        \textbf{Method}  & \textbf{Iris} & \textbf{Yacht} & \textbf{Housing} & \textbf{Diabetes} & \textbf{Blood} & \textbf{Energy} & \textbf{German} & \textbf{Concrete} & \textbf{Yeast} \\ 
        \midrule
        mean & .2634$\pm$.0054 & .2970$\pm$.0072 & .2471$\pm$.0140 & .4595$\pm$.0134 & .1654$\pm$.0036 & .3535$\pm$.0019 & .3163$\pm$.0185 & .2255$\pm$.0017 & .1185$\pm$.0009\\
        KNN & .1428$\pm$.0057 & .2630$\pm$.0057 & \runup{.1352$\pm$.0123} & .3339$\pm$.0084 & .1287$\pm$.0033 & .2408$\pm$.0022 & \runup{.2968$\pm$.0217} & .1746$\pm$.0032 & .1164$\pm$.0036\\
        MF & .1428$\pm$.0139 & .2472$\pm$.0086 & .1458$\pm$.0150 & .4579$\pm$.0134 & .1201$\pm$.0047 & .2291$\pm$.0039 & .3106$\pm$.0261 & .1776$\pm$.0032 & .1122$\pm$.0011\\
        MICE & .1454$\pm$.0142 & .2805$\pm$.0080 & .1852$\pm$.0162 & .4470$\pm$.0144 & .1337$\pm$.0045 & .2606$\pm$.0024 & .3374$\pm$.0207 & .2004$\pm$.0018 & .1286$\pm$.0003\\
        \midrule
        OT & .1393$\pm$.0108 & .2742$\pm$.0051 & .1657$\pm$.0174 & .3255$\pm$.0096 & .1442$\pm$.0033 & .2466$\pm$.0025 & .3078$\pm$.0197 & .1747$\pm$.0043 & .1164$\pm$.0008\\
        MIRACLE & .1364$\pm$.0183 & .2560$\pm$.0084 & .1633$\pm$.0161 & .3573$\pm$.0488 & \runup{.1157$\pm$.0028} & .2294$\pm$.0023 & .3027$\pm$.0214 & .1719$\pm$.0028 & \runup{.1118$\pm$.0013}\\
        GRAPE & .1343$\pm$.0206 & .2450$\pm$.0104 & .1382$\pm$.0172 & \runup{.3187$\pm$.0097} & .1194$\pm$.0067 & .2365$\pm$.0071 & .2981$\pm$.0191 & .1380$\pm$.0064 & .1144$\pm$.0012\\
        IGRM & \runup{.1197$\pm$.0123} & \runup{.2391$\pm$.0077} & .1363$\pm$.0175 & .3235$\pm$.0127 & .1182$\pm$.0047 & \sota{.1863$\pm$.0049} & .3032$\pm$.0206 & \sota{.1240$\pm$.0038} & .1161$\pm$.0013\\
        \midrule
        MIWAE & .1323$\pm$.0160 & .2669$\pm$.0027 & .1572$\pm$.0111 & .3750$\pm$.0223 & .1280$\pm$.0047 & .2528$\pm$.0012 & .3515$\pm$.0185 & .1894$\pm$.0068 & .1236$\pm$.0012\\
        GAIN & .1353$\pm$.0131 & .2526$\pm$.0082 & .1601$\pm$.0167 & .3942$\pm$.0166 & .1537$\pm$.0099 & .2702$\pm$.0097 & .3242$\pm$.0199 & .2203$\pm$.0021 & .1241$\pm$.0033\\
        TabCSDI & .1365$\pm$.0058 & .2588$\pm$.0062 & .1577$\pm$.0153 & .3636$\pm$.0132 & .1374$\pm$.0058 & .2611$\pm$.0047 & .2997$\pm$.0223 & .2477$\pm$.0036 & .1198$\pm$.0009\\
        FD & .1392$\pm$.0058 & .2509$\pm$.0081 & .1592$\pm$.0106 & .3789$\pm$.0044 & .1346$\pm$.0040 & .2592$\pm$.0053 & .3024$\pm$.0309 & .1927$\pm$.0016 & .1192$\pm$.0017\\
        \midrule
        \ourmethod & \sota{.1083$\pm$.0087} & \sota{.2324$\pm$.0141} & \sota{.1338$\pm$.0119} & \sota{.2937$\pm$.0031} & \sota{.1088$\pm$.0046} & \runup{.2194$\pm$.0053} & \sota{.2889$\pm$.0194} & \runup{.1366$\pm$.0008} & \sota{.1107$\pm$.0010}\\

        \bottomrule
        \toprule

        \textbf{Method}  & \textbf{Airfoil} & \textbf{Wine-red} & \textbf{Abalone} & \textbf{Wine-white} & \textbf{Phoneme} & \textbf{Power} & \textbf{Ecommerce} & \textbf{California} & \textbf{Average Rank} \\
        \midrule
        mean & .2898$\pm$.0024 & .1298$\pm$.0060 & .1952$\pm$.0011 & .1051$\pm$.0087 & .1622$\pm$.0007 & .1966$\pm$.0007 & .3193$\pm$.0066 & .1463$\pm$.0003 & 11.7\\
        KNN & .2513$\pm$.0018 & .0931$\pm$.0013 & .1305$\pm$.0006 & \runup{.0811$\pm$.0042} & .1344$\pm$.0009 & .1571$\pm$.0015 & \sota{.2959$\pm$.0113} & .1419$\pm$.0004 & 5.2\\
        MF & .2379$\pm$.0023 & .0980$\pm$.0025 & .1461$\pm$.0021 & .0858$\pm$.0066 & .1369$\pm$.0010 & .1512$\pm$.0012 & .3654$\pm$.0105 & .1186$\pm$.0005 & 6.1\\
        MICE & .2929$\pm$.0015 & .1071$\pm$.0028 & .1420$\pm$.0010 & .0920$\pm$.0076 & .1780$\pm$.0006 & .1631$\pm$.0014 & .3606$\pm$.0079 & .1292$\pm$.0007 & 10.2\\
        \midrule
        OT & .2726$\pm$.0039 & .0998$\pm$.0040 & .1499$\pm$.0018 & .0865$\pm$.0070 & .1573$\pm$.0008 & .1871$\pm$.0011 & .3177$\pm$.0067 & .1426$\pm$.0004 & 8.2\\
        MIRACLE & .2577$\pm$.0024 & .0942$\pm$.0027 & \runup{.1293$\pm$.0012} & .0865$\pm$.0070 & .1559$\pm$.0006 & .1442$\pm$.0010 & .3142$\pm$.0056 & .1175$\pm$.0005 & 4.9\\
        GRAPE & .2289$\pm$.0045 & .0897$\pm$.0033 & .1314$\pm$.0061 & .0982$\pm$.0198 & .1237$\pm$.0012 & \runup{.1338$\pm$.0016} & .3071$\pm$.0082 & \sota{.0997$\pm$.0006} & 3.5\\
        IGRM & \sota{.1797$\pm$.0033} & \sota{.0860$\pm$.0027} & .1358$\pm$.0063 & .0981$\pm$.0206 & \runup{.1228$\pm$.0011} & OOM & OOM & OOM & 3.2\\
        \midrule
        MIWAE & .2517$\pm$.0020 & .1138$\pm$.0000 & .1302$\pm$.0023 & .0864$\pm$.0074 & .1633$\pm$.0015 & .1578$\pm$.0028 & .3405$\pm$.0104 & .1285$\pm$.0005 & 7.6\\
        GAIN & .2581$\pm$.0019 & .1140$\pm$.0044 & .1493$\pm$.0113 & .1024$\pm$.0076 & .1685$\pm$.0027 & .1626$\pm$.0050 & .3450$\pm$.0119 & .1311$\pm$.0007 & 9.9\\
        TabCSDI & .2473$\pm$.0026 & .0994$\pm$.0078 & .1487$\pm$.0035 & .0952$\pm$.0079 & .1600$\pm$.0015 & .1526$\pm$.0020 & .3250$\pm$.0112 & .1287$\pm$.0010 & 7.8\\
        FD & .2458$\pm$.0020 & .1026$\pm$.0026 & .1468$\pm$.0013 & .0976$\pm$.0118 & .1596$\pm$.0014 & .1633$\pm$.0011 & .3357$\pm$.0087 & .1289$\pm$.0004 & 7.7\\
        \midrule
        \ourmethod & \runup{.2024$\pm$.0046} & \runup{.0873$\pm$.0016} & \sota{.1204$\pm$.0012} & \sota{.0727$\pm$.0056} & \sota{.1165$\pm$.0004} & \sota{.1320$\pm$.0009} & \runup{.3051$\pm$.0111} & \runup{.1056$\pm$.0006} & 1.4\\                  
        \bottomrule
    \end{tabular}
    }
    \label{tab:main}
\end{table*}

On the basis of Gaussian noise-based data perturbation, we propose a state-dependent data augmentation. The core idea is to perturb entries in different states with different strengths (i.e., Gaussian variance). Concretely, given an input vector of \ourmethod, each entry can be in three ``states'': ground-truth state (GT) where $\mathbf{m}_c=0$, pseudo missing state (PM) where $\mathbf{m}_p=0$, and missing state (MS) where $\mathbf{m}=1$. For entries in different states, their certainties can be quite different: for a GT entry, the data is fully reliable; for a PM entry, its certainty is moderate, since this entry is generated by adding ground-truth data with $t$-related noise; for a MS entry, it has the lowest certainty because we know nothing about the truth value of this entry. In our state-dependent data augmentation strategy, we propose to define the perturbation strength $p$ according to states. Concretely, for entries with lower certainty, we will allocate a higher perturbation strength to it. Formally, in training phase, Eq.~(\ref{eq:hybrid}) can be rewritten by:
\begin{equation}
\label{eq:hybrid_aug}
\begin{aligned}
&\tilde{\mathbf{x}}_t = {\mathbf{x}}_0 \odot \left(\mathbbm{1}_{d}-\mathbf{m}_c\right) + \mathbf{x}_t \odot \mathbf{m}_c + \boldsymbol{\xi} \odot \mathbf{p}, \\
&\mathbf{p}=\left(\mathbbm{1}_{d}-\mathbf{m}\right) \times p_{GT} + \mathbf{m}_p \times p_{PM} + \mathbf{m} \times p_{MS},
\end{aligned}
\end{equation}

\noindent where $\boldsymbol{\xi}$ is a zero-centered perturbation noise vector, $\mathbf{p}$ is the perturbation strength vector, $p_{GT}$, $p_{PM}$, and $p_{MS}$ are the perturbation strengths for the corresponding states that satisfy $p_{GT} < p_{PM} < p_{MS}$. With state-dependent augmentation, we can generate more reliable training samples for model training without disrupting the original data distribution. As a result, the model robustness can be boosted; Also, data augmentation can further facilitate self-supervised learning by amplifying the diversity within the data across two channels. The overall algorithm and complexity analysis of \ourmethod is demonstrated in Appendix~\ref{app:algo} and \ref{app:complex}, respectively.

\subsection{Extending to Mixed-Type Data}\label{subsec:mix}

Previous subsections introduce the Gaussian diffusion version of \ourmethod for pure numerical tabular data. In practice, we can easily extend \ourmethod to mixed-type data via the following simple modifications for categorical features.

At the diffusion model stage, we introduce multinomial diffusion process for categorical features, while keeping Gaussian diffusion process for numerical features. Similar to TabDDPM~\cite{kotelnikov2023tabddpm}, each categorical feature can be assigned a $K$-dimensional prediction head (where $K$ is the number of categories), followed by a Softmax function. The cross-entropy loss is used to optimize the categorical output. For self-supervised alignment, we minimize the output scores of each prediction head. For state-dependent data augmentation, we use random state transformation~\cite{multinomial_hoogeboom2021argmax} as the perturbation for categorical data. 

\section{Experiments}

\subsection{Experimental Settings}\label{subsec:setting}

\subsubsection{Dataset}
We evaluate the imputation performance on 17 real-world datasets across various domains from the UCI Machine Learning repository~\cite{asuncion2007uci} and Kaggle, including Iris, Yacht, Housing, Diabetes, Blood, Energy, German, Concrete, Yeast, Airfoil, Wine-red, Abalone, Wine-white, Phoneme, Power, Ecommerce, and California. We use the first two letters of each dataset name to indicate it for simplification purposes. The statistics of datasets are given in Appendix~\ref{app:dataset}.

\subsubsection{Baselines}
We compare \ourmethod with three groups of methods: 
1) shallow methods, including mean imputation, kNN imputation~\cite{knnimp_troyanskaya2001missing}, miss forest (MF)~\cite{stekhoven2012missforest}, and MICE~\cite{mice_van2011mice}; 2) deep methods, including OT~\cite{ot_muzellec2020missing}, MIRACLE~\cite{kyono2021miracle}, GRAPE~\cite{grape_you2020handling}, and IGRM~\cite{igrm_zhong2023data}; 3) deep generative methods, including MIWAE~\cite{mattei2019miwae}, GAIN~\cite{yoon2018gain}, TabCSDI~\cite{tabcsdi_zheng2022diffusion}, and ForestDiffusion (FD)~\cite{forestdiff_jolicoeur2023generating}. Among them, TabCSDI and FD are also based on diffusion models.

\subsubsection{Implementation Details}
In our major experiments, we simulate the default data missing setting as missing with complete random (MCAR) scenario with $30\%$ missing ratio. We use Root Mean Squared Error (RMSE) as our evaluation metric, which is commonly used in previous works~\cite{kyono2021miracle,ot_muzellec2020missing}. We report the averaged test accuracy over 5 runs of experiments. We reproduce baselines based on HyperImpute package~\cite{jarrett2022hyperimpute} or their corresponding official source codes. 
We select some important hyper-parameters through grid search, and keep the rest insensitive hyper-parameters to be fixed values. Concretely, the grid search is carried out on the following search space:
\begin{itemize}
    \item Diffusion steps: \{10, 50, 100\}
    \item Training epochs: \{10000, 20000, 30000\}
    \item Learning rate: \{0.001, 0.0001\}
    \item The number of layers: \{3,4,5\}
    \item Hidden dimensions: \{256, 512, 1024\}
    \item Trade-off parameter of self-supervised alignment loss $\gamma$: \{0.2, 0.5, 1, 3, 5, 10\}
    \item Perturbation strengths [$p_{GT}$,$ p_{PM}$,$ p_{MS}$]: \\
    \{ [0.0001,0.08,0.1], [0.0001,0.4,0.5], [0.001,0.8,1], [0.001,2,3] \}
\end{itemize}

\subsubsection{Computing Infrastructures}
We run all experiments with an Amazon Web Service (AWS) EC2 instance with the instance size \texttt{g4dn.xlarge}, which features a 4-core CPU, 16 GB Memory, and a Nvidia T4 GPU with 16 GB GPU Memory. We implement the proposed \ourmethod with Python 3.10 and PyTorch 1.11.0~\cite{paszke2019pytorch}.

\subsubsection{Source Code}
The source code of \ourmethod is available at: \url{https://github.com/yixinliu233/SimpDM}.

\subsection{Performance Comparison}

The performance comparison in the default setting is illustrated in Table~\ref{tab:main}. From the results, we have the following observations. 
\begin{enumerate}
    \item \ourmethod outperforms all the baselines on 11 datasets while achieving runner-up results on the rest 6 datasets. The consistent superiority of a majority of datasets highlights the adaptability and robustness of our method in addressing missing data imputation challenges. 
    \item Compared to diffusion model-based methods (i.e., TabCSDI and FD), \ourmethod has significantly better performance, indicating the effectiveness of self-supervised alignment and state-dependent augmentation.
    \item Among generative methods, two diffusion model-based approaches perform slightly better than the GAN/VAE-based methods. This observation demonstrates the potential of diffusion models in addressing imputation problems.
    \item Deep learning-based methods, such as MIRACLE, GRAPE, and IGRM, show competitive performance, indicating the superior capability of deep neural networks in data completion.
    \item In the realm of shallow methods, kNN and MF exhibit strong performance, underscoring their superior capability in tabular data imputation. 
    \item Compared to the strongest baseline IGRM, \ourmethod requires less memory, indicating the efficiency and scalability of our approach.
\end{enumerate}

\begin{figure}[tb]
	\centering
	\includegraphics[width=1\columnwidth]{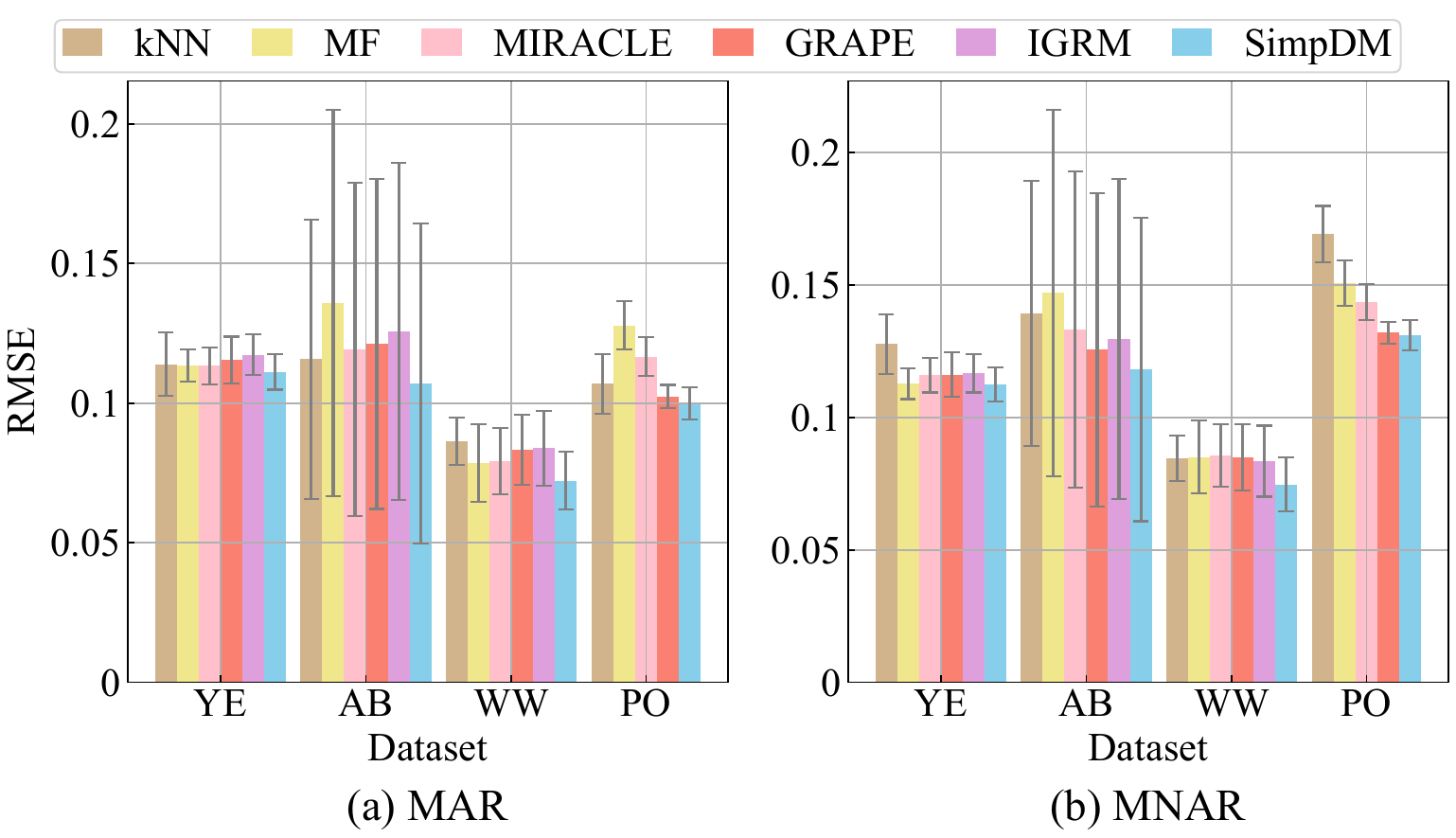} 
	\caption{Imputation performance on MAR and MNAR scenarios.}
	\label{fig:scenario}
\end{figure} 

\begin{figure}[tb]
	\centering
	\includegraphics[width=1\columnwidth]{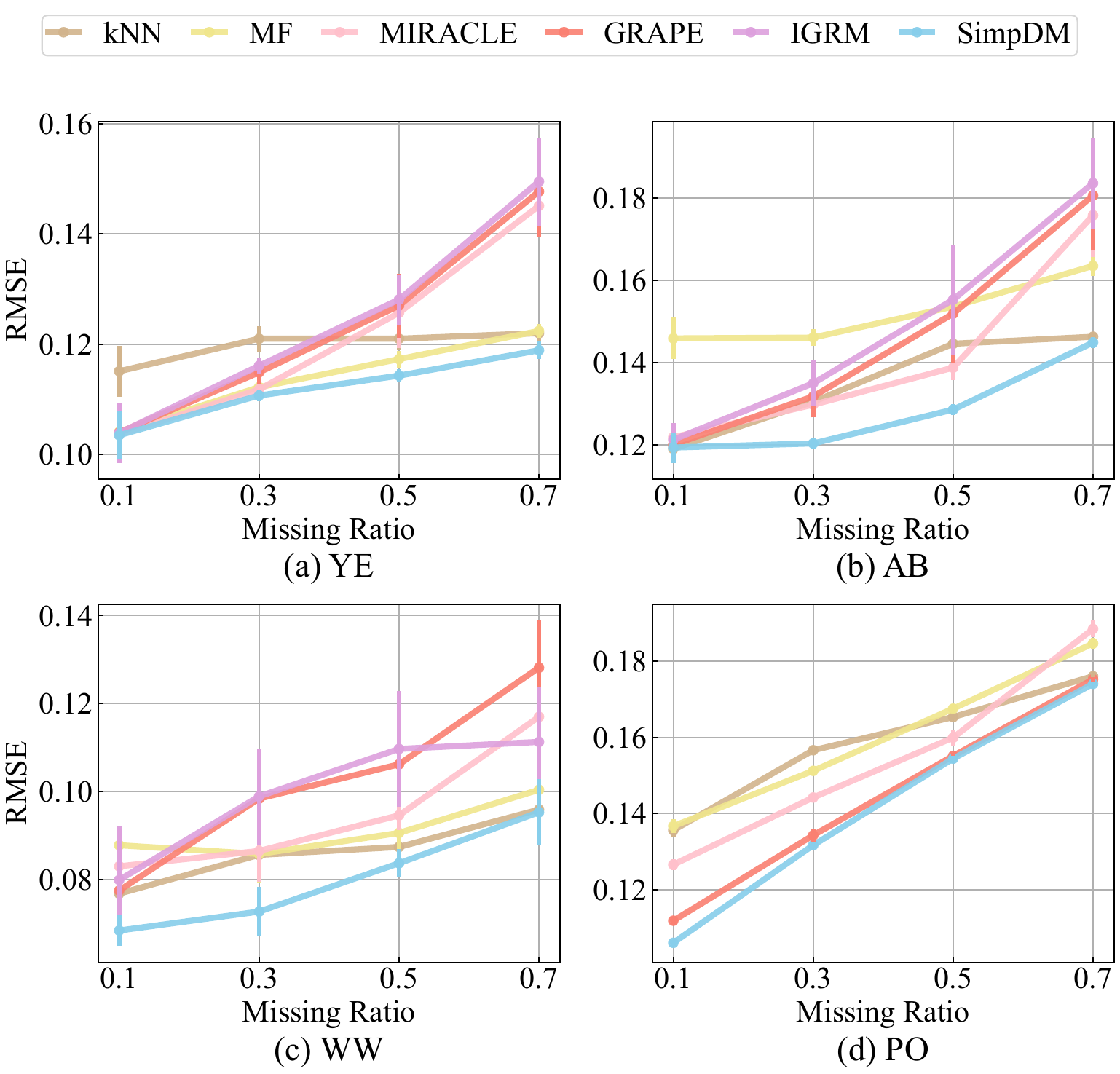} 
	\caption{Imputation performance under different missing ratios.}
	\label{fig:ratio}
\end{figure} 

\subsection{Different Missing Scenarios and Ratios}

To investigate the generalization ability in various missing situations, we conduct experiments on different missing scenarios (i.e., missing at random (MAR) and missing not at random (MNAR)) and missing ratios. For the sake of space, we compare \ourmethod with five highly competitive baselines, namely kNN, MF, MIRACLE, GRAPE, and IGRM. The experiments are conducted on four datasets, namely, Yeast (YE), Abalone (AB), Wine-white (WW), and Power (PO). 

\subsubsection{Different missing scenarios}
We conduct experiments to assess the effectiveness of \ourmethod in both MAR and MNAR scenarios, with results detailed in Fig.~\ref{fig:scenario}. The figures demonstrate that \ourmethod consistently surpasses all representative baselines in both scenarios, highlighting its robust generalization capabilities. Conversely, certain baselines exhibit suboptimal and unstable performance in certain cases.

\subsubsection{Different missing ratios}
To verify the robustness of \ourmethod across varying degrees of data missingness, we alter the missing ratios from $0.1$ to $0.7$, assessing its imputation performance across diverse scenarios. As depicted in Fig.~\ref{fig:ratio}, \ourmethod consistently demonstrates optimal or competitive performance across various missing ratios. Notably, in situations of severe data absence, \ourmethod exhibits more substantial performance gains compared to the deep baselines.

\subsection{Ablation Studies}\label{subsec:ablation}

\subsubsection{Effect of key components}
To examine the contributions of two key components in \ourmethod, namely self-supervised alignment (SA) and state-dependent augmentation (AUG), we conduct ablation experiments on our base model and its two variants, each incorporating one of these components. From the results in Table~\ref{tab:abl_main}, we can find that both SA and AUG yield substantial performance enhancements for the base model. Furthermore, SA demonstrates a more pronounced impact on performance across the majority of datasets. Finally, we can witness that \ourmethod, which incorporates both of these techniques, exhibits the best performance.

\begin{figure*}[t]
	\centering
        \subfigure[{Runtime Analysis}]{\includegraphics[height=0.5\columnwidth]{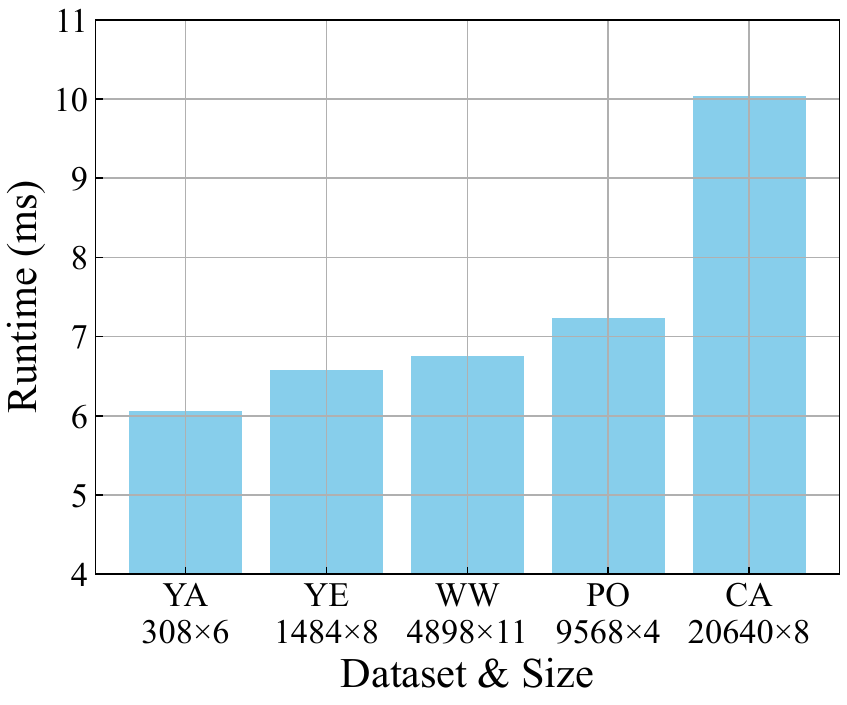}\label{fig:runtime}}
        \hfill
        \subfigure[{Case study}]{\includegraphics[height=0.5\columnwidth]{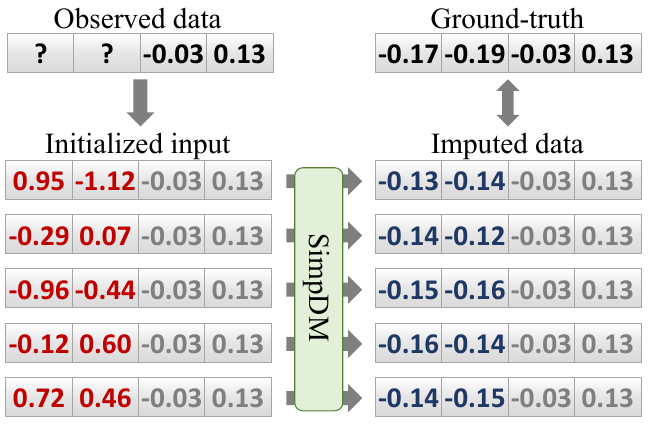}\label{subfig:case}}
        \hfill
	\subfigure[{Visualization}]{\includegraphics[height=0.5\columnwidth]{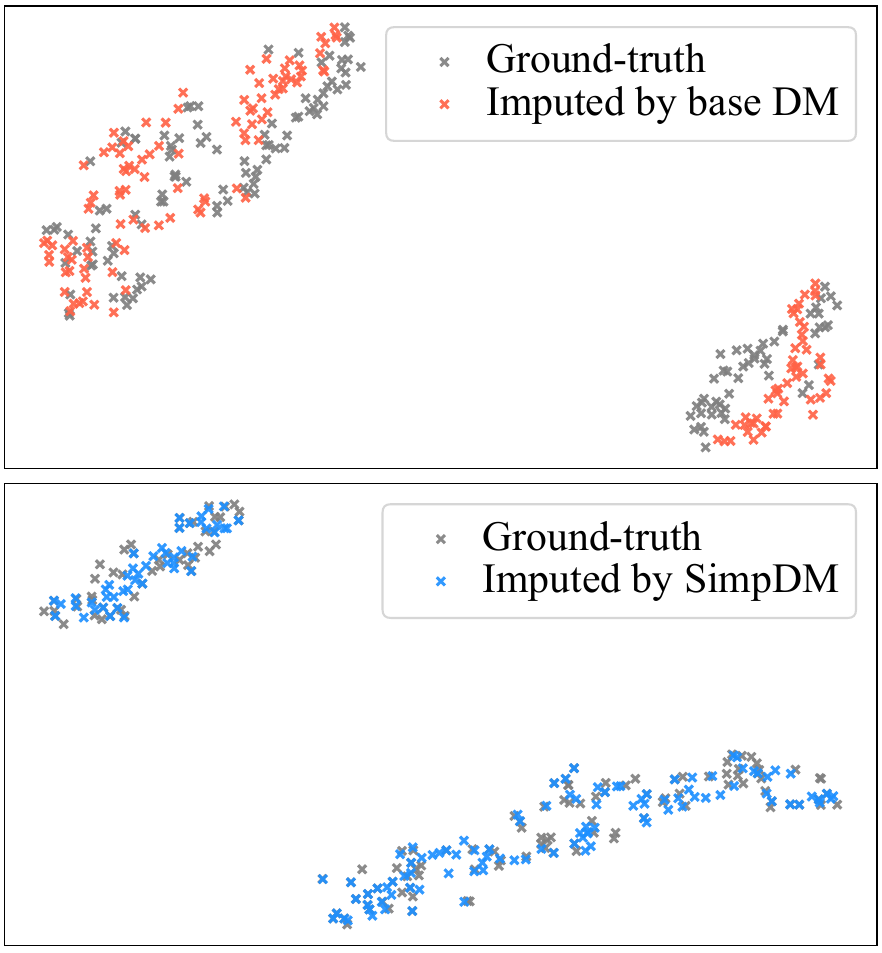}\label{subfig:vis}}
	\caption{(a) Runtime per training epoch on different datasets (with size $n \times d$). (b) The imputation results from different initialization. (c) t-SNE visualization of ground-truth data and imputed data.}
\end{figure*}

\subsubsection{Effect of different self-supervised losses}
In \ourmethod, self-supervised alignment loss is a critical component to eliminate the unexpected diversity of imputation results. We explore three self-supervised alignment loss types: MSE loss, contrastive loss (CL)~\cite{simclr}, and Sinkhorn divergence (SK)~\cite{ot_muzellec2020missing} (refer to Appendix xx for definitions), evaluating their respective effectiveness. As illustrated in Table~\ref{tab:abl_SA}, MSE loss attains the optimal results on 7 out of 8 datasets and exhibits competitive performance on the remaining one. We attribute the superior performance of MSE loss to its consistency with the objective of aligning the imputation results of two channels. Simultaneously, CL and SK losses also yield improvements, underscoring the effectiveness of self-supervised alignment. 

\begin{table}[t!]
\caption{Ablation study for the key components in \ourmethod.}
    \centering
    \resizebox{1.0\columnwidth}{!}{
    \begin{tabular}{l|cccccccc}
        \toprule
        \textbf{Variant}  & \textbf{YA} & \textbf{DI}  & \textbf{EN}  & \textbf{YE} & \textbf{WR} & \textbf{AB} & \textbf{WW} & \textbf{PO} \\

        \midrule
        Base model & .294 & .307 & .286 & .149 & .100 & .163 & .089 & .169 \\
        {+SA} & .243 & .298 & \runup{.232} & \runup{.118} & \runup{.091} & \runup{.130} & \runup{.076} & \runup{.140} \\
        {+AUG} & \runup{.241} & \runup{.297} & .241 & .123 & .092 & .132 & .077 & .149 \\
        \midrule
        \ourmethod & \sota{.232} & \sota{.294} & \sota{.219} & \sota{.111} & \sota{.087} & \sota{.120} & \sota{.073} & \sota{.132} \\
        \bottomrule
    \end{tabular}
    }
    \label{tab:abl_main}
\end{table}

\begin{table}[t!]
    \caption{Ablation study for self-supervised losses.}
    \centering
    \resizebox{1.0\columnwidth}{!}{
    \begin{tabular}{l|cccccccc}
        \toprule
        \textbf{Variant}  & \textbf{YA} & \textbf{DI}  & \textbf{EN}  & \textbf{YE} & \textbf{WR} & \textbf{AB} & \textbf{WW} & \textbf{PO} \\

        \midrule
        Base model & .294 & .307 & .286 & .149 & .100 & .163 & .089 & .169 \\
        +SA (CL) & .261 & .307 & \runup{.251} & \runup{.121} & .095 & \runup{.138} & .081 & \runup{.151} \\
        +SA (SK) & \runup{.259} & \sota{.295} & .259 & .134 & \runup{.092} & .148 & \runup{.080} & .154 \\
        +SA (MSE) & \sota{.243} & \runup{.298} & \sota{.232} & \sota{.118} & \sota{.091} & \sota{.130} & \sota{.076} & \sota{.140} \\
        \bottomrule
    \end{tabular}
    }
    \label{tab:abl_SA}
\end{table}

\begin{table}[t!]
    \caption{Ablation study for augmentation strategies.}
    \centering
    \resizebox{1.0\columnwidth}{!}{
    \begin{tabular}{l|cccccccc}
        \toprule
        \textbf{Variant}  & \textbf{YA} & \textbf{DI}  & \textbf{EN}  & \textbf{YE} & \textbf{WR} & \textbf{AB} & \textbf{WW} & \textbf{PO} \\

        \midrule
        Base model & \runup{.294} & .307 & .286 & .149 & .100 & \runup{.163} & \runup{.089} & \runup{.169} \\
        +AUG (strong) & .298 & .384 & \runup{.280} & \runup{.139} & .127 & .172 & .099 & .184 \\
        +AUG (weak) & .296 & \runup{.306} & .290 & .149 & \runup{.099} & .165 & .090 & .171 \\
        +AUG (SD) & \sota{.241} & \sota{.297} & \sota{.241} & \sota{.123} & \sota{.092} & \sota{.132} & \sota{.077} & \sota{.149} \\
        \bottomrule
    \end{tabular}
    }
    \label{tab:abl_aug}
\end{table}

\subsubsection{Effect of different augmentation strategies}
To verify the effectiveness of state-dependent (SD) augmentation, we compare it with two plain data augmentation strategies that perturb all entries with strong or weak strengths. From Table~\ref{tab:abl_aug}, we can see that weak data augmentation does not deeply affect the performance, whereas strong data augmentation tends to have a negative impact since it distorts the distribution of the original data. Conversely, state-dependent augmentation brings improvement by applying customized perturbation strengths to each entry according to its state and certainty.

\subsection{Runtime Analysis}

We present the runtime of each training epoch of \ourmethod on different datasets. All the experiments are conducted on an Amazon EC2 server (see Sec.~\ref{subsec:setting}) and with a fixed-size MLP diffusion model (with 3 layers and 256 hidden units). The experimental results are illustrated in Fig.~\ref{fig:runtime}. The figure clearly illustrates that the training time of \ourmethod exhibits a linear relationship with both the number of samples $n$ and the number of dimensions $d$. This observation aligns seamlessly with our analysis detailed in Appendix~\ref{app:complex}. Also, the training cost of \ourmethod is quite small (with few milliseconds for each epoch), indicating the high running efficiency and scalability of our method.

\subsection{Qualitative Analysis}

\subsubsection{Case study}
We perform a case study experiment on a sample from the Power dataset (the same one as Fig.~\ref{subfig:moti_target}) to investigate whether \ourmethod can deliver stable imputation results from diverse initial noises. From Fig.~\ref{subfig:case}, it is evident that \ourmethod yields both increased stability and enhanced accuracy. This improvement can be attributed to the integration of the self-supervised alignment mechanism.

\subsubsection{Visualization}
Using t-SNE algorithm~\cite{tsne_van2008visualizing}, we visualize the distribution of the original Iris dataset and the data imputed by the base model and \ourmethod, respectively. In Fig.~\ref{subfig:vis}, we observe a significant overlap between the distribution of data imputed by \ourmethod and the original data distribution, which suggests that \ourmethod adeptly captures the data manifold. In contrast, the base diffusion model struggles to match the data distribution.

\section{Conclusion}
In this paper, we introduce a novel variant of diffusion models, termed \ourmethod, designed for tabular data imputation. To enhance the imputation capabilities of the diffusion model, we propose a self-supervised alignment mechanism aimed at reducing its sensitivity to noise, thereby improving the stability of imputation results. Simultaneously, to address discrepancies in data scale, we present a state-dependent augmentation strategy that generates synthetic data during model training, which aims to bolster the robustness of \ourmethod. Extensive experiments showcase the superior performance of \ourmethod across various scenarios. 


\appendix

\section{Algorithm of \ourmethod} \label{app:algo}

\begin{algorithm}[t]
    \caption{\ourmethod Training}
    \label{alg:train}
    \textbf{Input}: Sample $\mathbf{x}$, missing mask $\mathbf{m}$ 
    \begin{algorithmic}[1] 
        \STATE $\mathbf{m}_p \sim \operatorname{Bernoulli} (\left(\mathbbm{1}_{n} - \mathbf{m} \right) \times r_m)$
        \STATE $\mathbf{m}_c = \mathbf{m}_p + \mathbf{m}$
        \FOR{$k = 1,2$} 
            \STATE $t_k \sim \operatorname{Uniform}(\{0, \ldots, T\})$
            \STATE $\epsilon_k \sim \mathcal{N}(0, \boldsymbol{I})$ 
            \STATE $\xi_k \sim \mathcal{N}(0, \boldsymbol{I})$
            \STATE Calculate $\tilde{\mathbf{x}}_{t_k}$ via Eq.~(\ref{eq:hybrid_aug})
            \STATE $\hat{\mathbf{x}}_k=\mathbf{x}_\theta\left(\tilde{\mathbf{x}}_{t_k}, {t_k}, \mathbf{m}_{c}\right)$
            \STATE Calculate $\mathcal{L}_{dm_k}$ via Eq.~(\ref{eq:ourdmloss})
        \ENDFOR
        \STATE{Calculate $\mathcal{L}_{sa} = \operatorname{MSE}(\hat{\mathbf{x}}_1,\hat{\mathbf{x}}_2)$}
        \STATE{Calculate overall loss $\mathcal{L}$ via Eq.~(\ref{eq:overall_loss}})
        \STATE Update model $\mathbf{x}_\theta$ by taking gradient descent step on $\mathcal{L}$
    \end{algorithmic}
\end{algorithm}

\begin{algorithm}[t]
    \caption{\ourmethod Imputing}
    \label{alg:test}
    \textbf{Input}: Sample $\mathbf{x}$, missing mask $\mathbf{m}$ 
    \begin{algorithmic}[1] 
        \STATE $\mathbf{x}_T \sim \mathcal{N}(0, \boldsymbol{I})$
        \FOR{$t = T, \cdots, 1$} 
            \STATE $\mathbf{z} \sim \mathcal{N}(\mathbf{0}, \mathbf{I})$ if $t>1$, else $\mathbf{z}=\mathbf{0}$
            \STATE Calculate $\tilde{\mathbf{x}}_{t}$ via Eq.~(\ref{eq:hybrid})
            \STATE $\mathbf{x}_{t-1}=\frac{1}{\sqrt{\alpha_t}}\big(\mathbf{x}_\theta\left(\tilde{\mathbf{x}}_{t}, {t}, \mathbf{m}\right)\big)+\sigma_t \mathbf{z}$
        \ENDFOR
        \STATE $\hat{\mathbf{x}}=\mathbf{x} \odot \left(\mathbbm{1}_{n \times d}-\mathbf{m}\right)+{\mathbf{x}}_0 \odot \mathbf{m}$
        \RETURN $\hat{\mathbf{x}}$

    \end{algorithmic}
\end{algorithm}

\begin{table}[t!]
\caption{The statistics of datasets, including the number of samples, numerical features, and categorical features.}
\vspace{-2mm}
    \centering
    \resizebox{1.0\columnwidth}{!}{
    \begin{tabular}{l|ccc}
        \toprule
        \textbf{Dataset}  & \textbf{\#Samples} & \textbf{\#Numerical Feat}  & \textbf{\#Categorical Feat} \\

        \midrule
        Iris & 150 & 4 & 0\\
    Yacht & 308 & 6 & 0\\
    Housing & 506 & 12 & 1\\
    Diabetes & 520 & 1 & 15\\
    Blood & 748 & 4 & 0\\
    Energy & 767 & 8 & 0\\
    German & 1,000 & 7 & 13\\
    Concrete & 1,030 & 8 & 0\\
    Yeast & 1,484 & 8 & 0\\
    Airfoil & 1,503 & 5 & 0\\
    Wine-red & 1,599 & 11 & 0\\
    Abalone & 4,177 & 7 & 1\\
    Wine-white & 4,898 & 11 & 0\\
    Phoneme & 5,404 & 5 & 0\\
    Power & 9,568 & 4 & 0\\
    Ecommerce & 10,999 & 3 & 7\\
    California & 20,640 & 8 & 0\\
        \bottomrule
    \end{tabular}
    }
    \label{tab:dset}
\end{table}

In this section, we provide the training and inference algorithms of \ourmethod with the example of a single data. In practice, we employ mini-batch-based training. Here we use Gaussian noise as the example of perturbation and employ MSE loss as the self-supervised alignment loss.  

The training algorithm of \ourmethod is summarized in Algorithm~\ref{alg:train}. In the first step, we  sample the pseudo mask $\mathbf{m}_p$ for the missing training strategy. The condition mask can be computed by adding $\mathbf{m}_p$ and the original missing mask $\mathbf{m}$. After that, we run the diffusion model on two channels ($k=1,2$) respectively. In each channel, we sample the diffusion time step $t_k$ and noise $\epsilon_k$ first. Then, we conduct the state-dependent augmentation with an extra sampled perturbation $\xi_k$ (Line 6-7). With the perturbed input, we obtain the imputed data with the diffusion model (Line 8) and calculate the diffusion model loss $\mathcal{L}_{{dm}_k}$ accordingly (Line 9). Once we obtain the imputed data by two channels ($\hat{\mathbf{x}}_1$ and $\hat{\mathbf{x}}_2$), we calculate the self-supervised alignment loss $\mathcal{L}_{sa}$. Finally, we can add the losses together and train the diffusion model $\mathbf{x}_\theta$ via gradient descent.

The inference (imputation) of \ourmethod is demonstrated in Algorithm~\ref{alg:test}. Similar to the vanilla diffusion model, we first initialize $\mathbf{x}_T$ with random Gaussian noise, and then recursively run the denoising iteration with the diffusion model. Differently, in each iteration, we set the observed entries to the ground-truth values (Line 4) before each denoising step, which ensures the observed values can well guide the imputation process. Finally, the imputed data $\hat{\mathbf{x}}$ can be obtain by combining $\mathbf{x}$ and the final estimation $\mathbf{x}_0$ (Line 7). 

\section{Complexity Analysis} \label{app:complex}

We study the time complexity of a single training epoch of \ourmethod, given a tabular dataset with $n$ samples and $d$ dimensions. For the sampling of pseudo mask, Gaussian noise, and augmented perturbation, their complexities are all $\mathcal{O}(nd)$. For the diffusion model, the time complexity is $\mathcal{O}(nd_h(d+d_hL))$, where $d_h$ and $L$ are the latent dimensions and the layer number of MLP, respectively. For the diffusion model loss and the MSE self-supervised alignment loss, the complexities are also $\mathcal{O}(nd)$. Note that if we use more complex self-supervised alignment loss (e.g., contrastive loss or Sinkhorn loss), the complexity can be higher to $\mathcal{O}(n^2d)$, which severely reduces the running efficiency. After eliminating the smaller terms, the training time complexity of \ourmethod becomes $\mathcal{O}(nd_h(d+d_hL))$. 
For the testing phase of \ourmethod, the complexity is also similar to vanilla diffusion models. In specific, the complexity is $\mathcal{O}(ndT)$ for the whole process, since we require $T$ diffusion time steps for data refinement. 
To sum up, this complexity of \ourmethod scales linearly with both $n$ and $d$, resembling the computational costs of vanilla diffusion models.

\section{Dataset} \label{app:dataset}

We conduct the experiments on 17 datasets from the UCI Machine Learning repository~\cite{asuncion2007uci} and Kaggle, including Iris, Yacht, Housing, Diabetes, Blood, Energy, German, Concrete, Yeast, Airfoil, Wine-red, Abalone, Wine-white, Phoneme, Power, Ecommerce, and California. The statistics of datasets are provided in Table~\ref{tab:dset}.

\clearpage
\bibliographystyle{ACM-Reference-Format}
\bibliography{ref}



\end{document}